\pdfoutput=1

\documentclass[11pt]{article}

\usepackage{acl}

\usepackage{times}
\usepackage{latexsym}
\usepackage{balance}
\usepackage[T1]{fontenc}
\usepackage{pgfplots} 
\usepackage{amssymb}
\usepackage[utf8]{inputenc}

\usepackage{multirow}
\usepackage{amsmath}
\usepackage{capt-of}
\usepackage{tabularx}
\usepackage{epsfig}
\usepackage{amssymb}
\usepackage{amsfonts}
\usepackage{booktabs}
\usepackage{scalerel}
\usepackage[inline]{enumitem}
\usepackage{listings}
\usepackage{varwidth}
\usepackage[export]{adjustbox}
\usepackage{tikz}
\usetikzlibrary{tikzmark}
\usepackage{cleveref}

\usepackage{stmaryrd}
\usepackage{bbm}

\usepackage{algorithm}
\usepackage[noend]{algpseudocode}

\definecolor{deepblue}{rgb}{0,0,0.5}
\definecolor{officeblue}{RGB}{0,102,204}
\definecolor{deepred}{rgb}{0.6,0,0}
\definecolor{deepgreen}{rgb}{0,0.5,0}
\definecolor{mybrickred}{RGB}{182,50,28}

\definecolor{fillcolor}{RGB}{216,217,252}

\algnewcommand\algorithmicrequireb{{\hspace{0.85cm}}}
\algnewcommand\INPTDESCB{\item[\algorithmicrequireb]}

\algnewcommand\algorithmicfuncdesc{\textbf{Function:}}
\algnewcommand\FUNCDESC{\item[\algorithmicfuncdesc]}
\algnewcommand\algorithmicfuncdescb{{\hspace{1.48cm}}}
\algnewcommand\FUNCDESCB{\item[\algorithmicfuncdescb]}
\algnewcommand{\algorithmicgoto}{\textbf{goto}}
\algnewcommand{\Goto}[1]{\algorithmicgoto~\ref{#1}}

\usepackage{amsmath,amsfonts,bm}

\def\eqref#1{equation~\ref{#1}}

\def\1{\bm{1}}

\def\vtheta{{\bm{\theta}}}
\def\va{{\bm{a}}}

\def\ve{{\bm{e}}}

\def\vg{{\bm{g}}}

\def\vs{{\bm{s}}}

\DeclareMathAlphabet{\mathsfit}{\encodingdefault}{\sfdefault}{m}{sl}
\SetMathAlphabet{\mathsfit}{bold}{\encodingdefault}{\sfdefault}{bx}{n}

\newcommand{\Ls}{\mathcal{L}}

\newcommand\eduSEP{\textsc{[EDU]}}
\newcommand\final{\textsc{FINAL}}

\newcommand\oursLong{ET5}
\newcommand\oursLongDiscern{ET5-Discern}

\title{ET5: A Novel End-to-end Framework for Conversational Machine Reading Comprehension}

\author{Xiao Zhang$^{123}$,~~Heyan Huang$^{123}$\thanks{\ \ Corresponding author.},~~Zewen Chi$^{123}$,~~\textbf{Xian-Ling Mao}$^{123}$\\
$^{1}$School of Computer Science and Technology, Beijing Institute of Technology\\
$^2$Beijing Engineering Research Center of High Volume Language Information Processing\\
and Cloud Computing Applications\\
$^3$Southeast Academy of Information Technology, Beijing Institute of Technology\\
\texttt{\{yotta,hhy63,czw,maoxl\}@bit.edu.cn}\\}

\usepackage{pgfplots}
\pgfplotsset{compat=1.17}

\begin{document}
\maketitle
\begin{abstract}
Conversational machine reading comprehension (CMRC) aims to assist computers to understand an natural language text and thereafter engage in a multi-turn conversation to answer questions related to the text.
Existing methods typically require three steps: (1) decision making based on entailment reasoning; (2) span extraction if required by the above decision; (3) question rephrasing based on the extracted span.
However, for nearly all these methods, the span extraction and question rephrasing steps cannot fully exploit the fine-grained entailment reasoning information in decision making step because of their relative independence, which will further enlarge the information gap between decision making and question phrasing. Thus, to tackle this problem, we propose a novel end-to-end framework for conversational machine reading comprehension based on shared parameter mechanism, called entailment reasoning T5 (\oursLong{}). Despite the lightweight of our proposed framework, experimental results show that the proposed \oursLong{} achieves new state-of-the-art results on the ShARC leaderboard with the BLEU-4 score of 55.2. Our model and code are publicly available\footnote{\url{https://github.com/Yottaxx/ET5}}.
\end{abstract}

\section{Introduction}
\begin{figure}[t]
\centering
\includegraphics[width=0.45\textwidth]{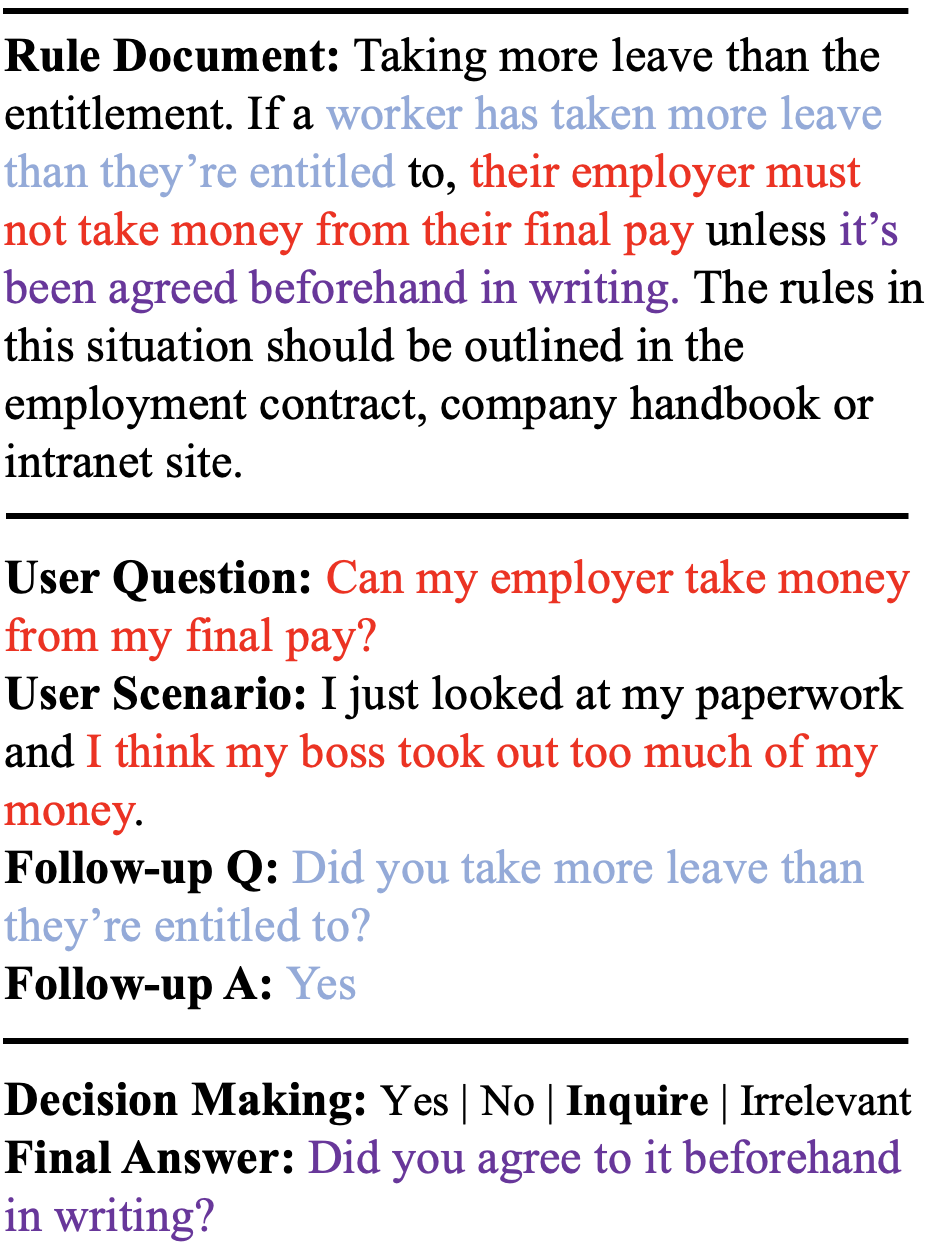}
\caption{An example in the CMRC dataset. Machine should first make the decision of \texttt{Yes/No/Inquire/Irrelevant}, and then generate the follow-up question if the decision is \texttt{Inquire}. 
The colored sentences show the reasoning process for the final answer. }
\label{fig:instance}
\end{figure}

Conversational machine reading comprehension (CMRC) \cite{sharc} aims to assist machines to understand an natural language text and thereafter engage in a multi-turn conversation to answer questions related to the text. Specifically, the machine needs to reason for decision making and question generation by interacting through rule document, user question, user scenario, and dialogue history. As an example shown in Figure \ref{fig:instance}, after fully interacting with complicated context information, the machine makes a decision of \texttt{Yes/No/Inquire/Irrelevant}, and then generates a question under the \texttt{Inquire} decision.


\begin{figure}[t]
\centering
\includegraphics[width=0.48\textwidth]{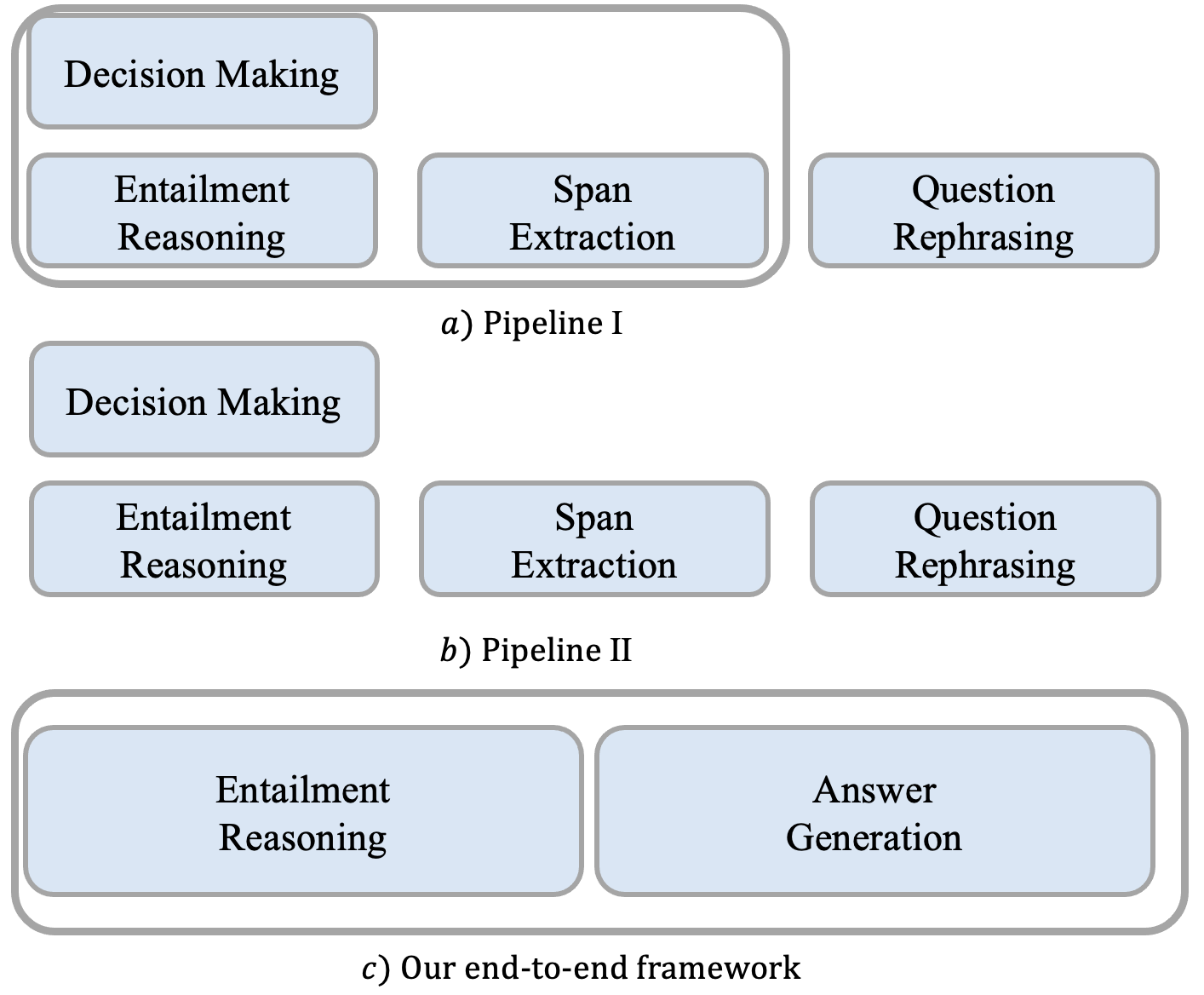}
\caption{The overview of frameworks in CMRC.
(a) For Pipeline I, decision making and span extraction models share the encoder but suffer from the problem of noisy span extraction. (b) For Pipeline II, the three stages are handled completely separately, and there is no information sharing among the three stages. (c)  Our framework is an end-to-end framework with a shared encoder and a duplex decoder. The duplex decoder contains an entailment reasoning decoder and answer generation decoder, both the information of entailment reasoning and answer generation are shared through the common encoder. Both decisions and follow-up questions will be generated via answer generation decoder directly.}
\label{fig:pipeline}
\end{figure}

Existing researches \cite{sharc,UrcaNet,lawrence-etal-2019-bison-stage1,zhong-zettlemoyer-2019-e3-span-generation,emt-span-generation,discern-span-generation,graph-span-generation} mainly aim to capture the interactions among the complicated inputs, and achieve promising results by conducting various fine-grained entailment reasoning interaction strategies based on Pre-trained Language Models (PrLMs) \cite{devlin-etal-2019-bert,liu2020roberta,dong2019unified,Clark2020ELECTRA:,t5}. These methods \cite{zhong-zettlemoyer-2019-e3-span-generation,emt-span-generation,discern-span-generation,graph-span-generation} typically adopt pipeline architectures, which are shown in Figure \ref{fig:pipeline}. These pipeline architectures typically require three steps : (1) decision making based on entailment reasoning; (2) span extraction if required by the above decision; (3) question rephrasing based on the extracted span. There are currently two types of pipeline structures: Pipeline I and Pipeline II. The Pipeline I make decisions and extract spans simultaneously, while the Pipeline II handles all three stages separately.

However, for nearly all these methods \cite{zhong-zettlemoyer-2019-e3-span-generation,emt-span-generation,discern-span-generation,graph-span-generation}, the span extraction and question rephrasing steps can’t fully exploit the fine-grained entailment reasoning information in decision making step. For Pipeline II, these methods \cite{discern-span-generation,graph-span-generation} do not share entailment reasoning information among decision-making, span extraction, and question phrasing at all. For Pipeline I, these methods \cite{zhong-zettlemoyer-2019-e3-span-generation,emt-span-generation} only approximate share the information through noisy span extraction. Both of them enlarge the information gap between decision making and question rephrasing, and seriously affect the performance of question generation.


To tackle this problem, we propose a novel end-to-end framework for conversational machine reading comprehension based on shared parameter mechanism, called entailment reasoning T5 (\oursLong{}).
Specifically, the proposed framework consists of a text-to-text Transformer and an additional entailment reasoning decoder. The original decoder in the text-to-text Transformer will directly generate either decision or follow-up question based on the shared encoder enhanced by entailment reasoning. The entailment reasoning decoder can be configured with different entailment reasoning strategies. Despite the lightweight of our proposed framework, experimental results show that \oursLong{} achieves new state-of-the-art results on the ShARC leaderboard with the BLEU-4 score of 55.2 and significantly improves the generalization performance of question generation.

Our contributions are summarized as follows:
\begin{itemize}
\item We propose a novel end-to-end framework, called  \oursLong{}, to better capture the entailment information for question generation, and thus eliminate the information gap between decision making and question generation.


\item Extensive experiments demonstrate the effectiveness of the proposed framework on ShARC benchmark, especially in the question generation sub-task.



\end{itemize}

\section{Related Work}


Conversation-based reading comprehension \cite{sharc,sun-etal-2019-dream,reddy-etal-2019-coqa,choi-etal-2018-quac,cui-etal-2020-mutual,gao-open-cmrc} extends the context with dialogue history, which is formed to simulate the communication scene in real life. Most of them are ideal subtasks, either span-based QA tasks \cite{choi-etal-2018-quac,reddy-etal-2019-coqa} or multi-choice tasks \cite{sun-etal-2019-dream,cui-etal-2020-mutual}. We focus on the task \cite{sharc} that deal with real-world complexities, where the machine needs to make decisions or ask questions to keep the conversation going. This task \cite{sharc} is called Conversational Machine Reading Comprehension (CMRC), 
which requires the machine to have the inference ability to capture the interactions among rule document, user question, user scenario, and dialogue history.

Recent studies \cite{zhong-zettlemoyer-2019-e3-span-generation,emt-span-generation,discern-span-generation,graph-span-generation} in CMRC are generally utilized to match the relationship between the various information. $\mathrm{E^3}$ \cite{zhong-zettlemoyer-2019-e3-span-generation} first investigates the importance of clarifying the different rule units for entailment reasoning.  Different entailment reasoning strategies \cite{emt-span-generation,discern-span-generation,gao-open-cmrc} with fine-grained reasoning units are further proposed to improve the abilities of entailment reasoning. In addition, discourse relationships between fine-grained reasoning units are utilized to model the discourse graph \cite{graph-span-generation,smoothing-open-cmrc}. These methods typically adopt pipeline architectures, DISCERN \cite{discern-span-generation} first discovers the unbalance and noisy problems of Pipeline I conducted by $\mathrm{E^3}$ \cite{zhong-zettlemoyer-2019-e3-span-generation} and EMT \cite{emt-span-generation}, then solves them by utilizing Pipeline II to process the three stages separately. However, due to the pipeline's inability to make full use of the entailment information, both of the above pipeline structures have the problem of information gap \cite{smoothing-open-cmrc} between decision making and question generation.

\begin{figure*}[t]
\centering
\includegraphics[width=0.99\textwidth]{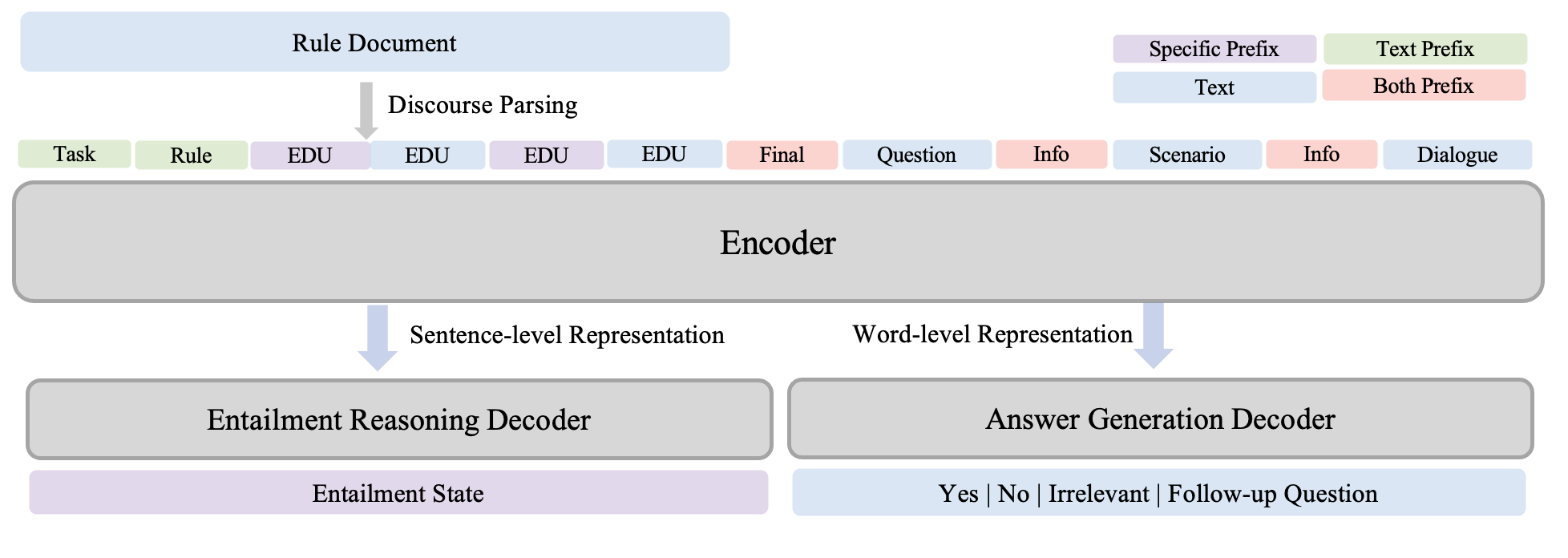}
\caption{The architecture of \oursLong{}. Our proposed framework is an end-to-end framework based on a single text-to-text Transformer. The decoder of our proposed framework is a duplex decoder, including entailment reasoning decoder and answer generation decoder. The answer generation decoder will generate the final answer directly, either of the decision or the follow-up question. The entailment reasoning decoder is utilized to reason the fine-grained entailment states, which is only activated in the training stage. Red boxes indicate the fine-grained prefixes, including special prefixes and text prefixes, which are represented by purple boxes and green boxes, respectively. Special prefixes refer to
 special tokens that aim to get the sentence-level representations. Text prefixes refer to the component-specific text prefixes that aim to differentiate among different input types.}
\label{fig:framework}
\end{figure*}

To better capture the entailment information for question generation and eliminate the information gap, we propose a novel end-to-end framework for conversational machine reading comprehension based on shared parameter mechanism, called \oursLong{}, which will be introduced in the next section.

\section{Method}

\subsection{Settings of \oursLong{}}

Each example of CMRC is formed as the tuple $\left \{ C,R,A,S \right \}$. $C$ donates the context, which is a concentrated sentence of rule document, user scenario, user question, and dialogue history. Especially, $C = \left \{ e_{1},e_{2},...,e_{k},s,q,d_{1},d_{2},...,d_{n} \right \}$,  where $e$ donates the elementary discourse unit (EDU) segmented from by rule documents.  $s$ and $q$ are user scenario and user question, $d$ represents the dialogues. 
Each item of $C$ is prefixed with a special token to represents the following sentence, the details of the prefix are written in Section 3.2. $R$ represents the discourse relations among EDUs, the parsed details are reported in Section 4.1. $A$ is the final answer, including the decision or follow-up question. $S$ donates the entailment reasoning state of each EDU in \texttt{ENTAILMENT}, \texttt{CONTRADICTION}, or \texttt{NEUTRAL}. To get the noisy supervision signals of entailment states, we adopt a heuristic approach\footnote{ The noisy supervision signal is a heuristic label obtained by the minimum edit distance.} following the previous study \cite{emt-span-generation}. Given inputs $C,R$,  \oursLong{} needs reasoning entailment states $S$ and final answer $A$ including the decision and follow-up question. As illustrated in Figure \ref{fig:framework}, we conduct duplex decoder to process answer generation and entailment reasoning simultaneously 
in a multi-task training approach with the shared encoder. The training procedure and evaluating procedure are illustrated in Algorithm~\ref{alg:trainP} and Algorithm~\ref{alg:evalP}, respectively. 

\begin{algorithm}[t]
\caption{Training procedure of \oursLong{}}
\label{alg:trainP}
\begin{algorithmic}[1]
\Require Concentrated context $C$, discourse relations $R$, learning rate $\tau$, discourse relations $R$
\Ensure Final answer $A$, entailment reasoning state $S$, \oursLong{} encoder parameters $\vtheta_{e}$, \oursLong{} answer generation decoder parameters $\vtheta_{a}$, \oursLong{} entailment reasoning decoder parameters $\vtheta_{d}$
\State Initialize $\vtheta_{e}$,$\vtheta_{a}$,$\vtheta_{d}$
\While{not converged}
\For {$i = 1,2,\dots,N$}
\State $\ve_i = f(c_i, \vtheta_{e}) $ ~~s.t. $\forall c \in \mathcal{C}$
\State $\vs_i = f(e_i,r_i,\vtheta_{d})$ ~~s.t. $\forall r \in \mathcal{R}$
\State $\va_i = f(e_i,\vtheta_{a})$

\EndFor
\State \textbf{end for}
\State $\vg \gets \nabla_\vtheta \Ls_{}$
\State $\vtheta_{e} \gets \vtheta_{e} - \tau \vg$
\State $\vtheta_{d} \gets \vtheta_{d} - \tau \vg$
\State $\vtheta_{a} \gets \vtheta_{a} - \tau \vg$

\EndWhile
\State \textbf{end while}
\end{algorithmic}
\end{algorithm}

\begin{algorithm}[t]
\caption{Evaluating procedure of \oursLong{}}
\label{alg:evalP}
\begin{algorithmic}[1]
\Require Concentrated context $C$, \oursLong{} encoder parameters $\vtheta_{e}$, \oursLong{} answer generation decoder parameters $\vtheta_{a}$
\Ensure Final answer $A$
\State Initialize $\vtheta_{e}$,$\vtheta_{a}$
\For {$i = 1,2,\dots,N$}
\State $\ve_i = f(c_i, \vtheta_{e}) $ ~~s.t. $\forall c \in \mathcal{C}$
\State $\va_i = f(e_i,\vtheta_{a})$
\EndFor
\State \textbf{end while}
\end{algorithmic}
\end{algorithm}

\subsection{Encoding}

\paragraph{Fine-grained Prefix Prompt}
 We investigate and propose a fine-grained prefix strategy, to prompt the interactions among different components of the input. As shown in Figure \ref{fig:framework}, the concatenate input is prefixed with a text-form task prefix. Furthermore, given relationship tagged EDUs, user question, user scenario, dialogue history as inputs, each of them is prefixed with a fine-grained prefix.The fine-grained prefix consists of a text prefix and a special prefix. The text prefix is used to differentiate between different types of input as an addition information prefix. The special prefix is used to obtain sentence-level representations required for entailment reasoning. In the case of user scenario and each dialogue history usually play a similar role as an information provider in CMRC tasks, user scenario and each dialogue history share the same text prefix. Meanwhile, each EDU has a special token \eduSEP{}. Both user question and final answer are prefixed with the same text \final{}. We concatenate the fine-grained prefixed EDUs, user question, user scenario, dialogue history to get the encoding representations.

\paragraph{Fine-grained Prefix Encoding}
We concatenate the fine-grained prefixed EDUs, user question, user scenario, dialogue history as the input. Encoder representation $H_{e}$ is encoded with the input by conducting T5 encoder \cite{t5} as the encoder. Let $ H_{s} = [h_{e_{1}}, h_{e_{2}}, ... , h_{e_{k}}, h_{f_{i}}, h_{s_{i}}, h_{d_{1}}, ..., h_{d_{n}}]
$, $H_{s} $ to donate the sentence-level representations. $h_{e}$, $h_{f}$, $h_{s}$, $h_{d}$  represent the fine-grained special prefix token representation of EDU, user question, user scenario, and dialogue history, respectively.


\subsection{Decoding}
Our decoding is duplex decoding, including entailment reasoning and answer generation. Especially, both answer generation and entailment reasoning are activated in the training stage. During the inference stage, the answer generation decoder will directly generate either the decision or the follow-up question while the entailment reasoning will be dropped. 
We conduct entailment reasoning decoder with various entailment reasoning strategies in the experiments, including inter attention reasoning \cite{discern-span-generation} and dialogue graph modeling \cite{graph-span-generation}. We mainly introduce dialogue graph reasoning here, because dialogue graph modeling only has one more graph reasoning block than inter attention reasoning, the other structures are the same.

\paragraph{Entailment Reasoning}
We utilize dialogue graph modeling for entailment reasoning decoding.  Dialogue graph consists of the explicit discourse graph, the implicit discourse graph, and the inter attention reasoning. The details are shown in the following.

Given $H_{s}$ and $R$, we construct the explicit discourse graph $G$ to explicitly model the complex logical structures between the various information in CMRC by introducing discourse relationships among the rule conditions. Following previous \cite{graph-span-generation}, the graph is formed as a Levi graph \cite{levi1942finite}. 

There are three types of vertices in the graph: EDUs, discourse relationships, and user scenarios. Each EDU duplex connects with the tagged relationship. The user scenario connects all the other vertices as a global vertex. All the types $ R_{L}$ of the possible edges between vertices are six, each of them is named as \textit{default-in}, \textit{default-out}, \textit{reverse-in},\textit{reverse-out}, \textit{self}, and \textit{global}. The EDUs vertices and user scenario vertex are initialized with the contextualized representation in $H_{s}$. And the discourse relationships vertices are initialized with a conventional embedding layer. Then the representation $h_{p}$ of each node $v_{p}$ is initialized. To handle the multi-relation graphs and dynamically weight the different relations, we use a relational graph convolution network \cite{schlichtkrull2018modeling-graph} with a gating mechanism. the graph-based information processing can be written as:
\begin{equation}
g_{p}^{(l)} = \mathrm{Sigmoid}(h_{p}^{(l)}w_{r,g}^{l}),
\end{equation}
\begin{equation}
h_{p}^{(l+1)} = \mathrm{ReLU} (\sum_{r \in R_{L}} \sum_{v_{p} \in \mathcal{N}_{r}(v_{p})}g_{p}^{(l)}\frac{1}{c_{p,r}}w_{r}^{(l)}h_{q}^{(l)}),
\end{equation}
where $w_{r}^{(l)}$ is the trainable parameters of layer $l$. $w_{r,g}^{(l)}$ is trainable parameters under relation type $r$ of layer $l$. $c_{p,r}$ is the number of the neighbors of node $v_{p}$ with relationship $r$. $\mathcal{N}_{r}(v_{p})$ refers to those neighbors. Let $ H_{p} = [h_{p_{1}}^{(l+1}), h_{p_{2}}^{(l+1)}, ... , h_{f_{i}}, h_{s_{i}}, h_{d_{1}}, ..., h_{d_{n}}]
$, $l$ is the last layer, $H_{p} $ donate the explicit discourse graph representation.

Given the EDUs tokens hidden representation $E$ from $H_{e}$.  We decouple and fuse the local information and the contextualized information by conducting the implicit discourse graph. Considering each token $i$ of EDU as a vertex in the graph, the adjacent matrices can express the implicit discourse graph. We use $I_{i}$ donate the index of token $i$ in EDU, the information decoupling adjacent matrices $M$ can be written as:

\begin{equation}
M_{l}[i,j] =
\begin{cases}
 \quad0  ,& \quad I_i = I_j\\ 
 -\infty, & otherwise
\end{cases}
\end{equation}

\begin{equation}
M_{c}[i,j] = \begin{cases}
 \quad0 ,& \quad I_i \neq  I_j\\ 
 -\infty , & otherwise
\end{cases},
\end{equation}
where $M_{l}$ and $M_{c}$ are conducted to express the local and contextualized information. We use multi-head-self-attention (MHSA) \cite{NIPSattention} to process decoupling:
\begin{equation}
G_{i} = \mathrm{MHSA}(E,M_i), \quad i \in \left \{ l,c \right \},
\end{equation}
after exploring the potential textual relations in the rule document, we apply a fusion layer to fuse the information by considering the encoder encoding and the attention hidden states of EDUs:

\begin{align}
\tilde{E_{1}} = \mathrm{ReLU}(f([E,G_{l},E-G_{l},E \odot G_{l} ])), \\
\tilde{E_{2}} = \mathrm{ReLU}(f([E,G_{c},E-G_{l},E \odot G_{c} ])), \\
g = \mathrm{Sigmoid}(f([\tilde{E_{1}},\tilde{E_{2}}])]), \\
C = g \odot G_{l} + (1-g) \odot G_{c},
\end{align}
where $f$ is the fully-connected layer. Let $ H_{i} = [h_{c_{1}}, h_{c_{2}}, ... , h_{f_{i}}, h_{s_{i}}, h_{d_{1}}, ..., h_{d_{n}}]
$, $H_{i} $ donate the explicit discourse graph representation. $h_{c_{i}}$ is updated by the representation of \eduSEP{} in $C$.

Given the sentence-level representation $H_{e}$, $H_{p}$,  $H_{i}$, inter attention reasoning aims to fully interact with various information, including EDUs, user question, user scenario, dialogue history. We utilize an inter-sentence Transformer \cite{NIPSattention} to reason the entailment states. Let $\tilde{H_{e}}$, $\tilde{H_{p}}$,  $\tilde{H_{i}}$ donate the inter-sentence Transformer encoding representation,  $\tilde{H}_{s}$ donate the average encoding, namely, $\tilde{H}_{s} = [\tilde{h}_{e_{1}}, \tilde{h}_{e_{2}}, ... , \tilde{h}_{e_{k}}, \tilde{h}_{f_{i}}, \tilde{h}_{s_{i}}, \tilde{h}_{d_{1}}, ..., \tilde{h}_{d_{n}}]
$. All the vectored representations are in the same dimension. Following previous studies \cite{gao-open-cmrc}, we utilize a linear transformation to track the entailment reasoning state of each EDU:
\begin{equation}
c_{i} = W_{c} \tilde{h}_{e_i} + b_{c} \in \mathcal{R}^{3},
\end{equation}
where the $W_c$ is trainable parameters, $c_{i}$ is the predicted score for the three labels of the $i$-th states.

\paragraph{Answer Generation}

Answer generation is utilized to generate either the decision or the follow-up question. We employ T5 decoder as our answer generation decoder. Given encoder hidden representation $H_{e}$, and the set of final answer $(a_{1},a_{2},...,a_{n})$,  including decision or follow-up question, each of the answers is composed of the variable-length tokens $(x_{1},x_{2},...,x_{m})$, the probabilities over the tokens are shown in the blow:

    \begin{equation}
        p(a) = \prod_{1}^{m}p(x_{i}|x_{<i},H_{e};\theta ),
    \end{equation}
where $\theta$ donates the trainable parameters of our decoder.

\subsection{Training Objective}

\paragraph{Entailment Reasoning}
 Given the entailment fulfillment states $c_{i}$, the entailment reasoning is supervised by cross-entropy loss:
\begin{equation}
    \mathcal{L}_{enatil} = - \frac{1}{N} \sum_{i=1}^{N} log \, \mathrm{softmax} (c_i)_{r},
\end{equation}
where $r$ is the ground truth of entailment state.

\paragraph{Answer Generation}
Given the encoder representation $H_{e}$, the answer generation training objective is computed by:
    \begin{equation}
        \mathcal{L}_{answer} = -\sum_{i=1}^{M} log \, p(x_{i}|x_{<i},H_{e};\theta ),
    \end{equation}
The overall loss function is:
\begin{equation}
    \mathcal{L} = \mathcal{L}_{answer} + \lambda \mathcal{L}_{entail}.
\end{equation}




\section{Experiment and Analysis}

\subsection{Data}
\paragraph{Dataset}
The experimental dataset is ShARC, the current CMRC benchmark, which is built up from 948 rule text. The corpus is clawed from the government website. The utterances size of ShARC is 32,436, each of the utterances related to a dialog tree, the utterances with the same rule text refer to the same dialog tree. Each dialog tree contains all possible fulfillment combinations of conditions. The train, dev, test size is 21,890, 2,270, 8,276, respectively. Each item consists of utterance id, tree id, rule document,
initial question, user scenario, dialog history, evidence, and the decision. Evidence is only used to support the answer, and can't be treated as input.

\paragraph{Preprocess}
Following previous methods \cite{graph-span-generation,discern-span-generation}, we first split rule documents into elementary discourse units (EDUs), and then tag the discourse relationship among EDUs. For discourse segmentation, the rule documents are split into EDUs by using a pre-trained discourse parser \cite{li2018segbotParser}. For discourse relation extraction, we utilize a pre-trained discourse relation parser\footnote{\url{https://github.com/shizhouxing/DialogueDiscourseParsing}} to tag the structural relations 
among EDUs. 



\subsection{Setup}

\paragraph{Evaluation}
Evaluation in ShARC is divided into two parts. First is decision classification: Micro-Acc and Macro-Acc scores are used for the evaluation in classification. Then question generation part is evaluated with BLEU \cite{papineni-etal-2002-bleu} score only if the prediction and ground truth in classification are both inquired.

\paragraph{Implementation Details}
We implement \oursLong{} by configuring entailment reasoning decoder with two different methods: inter attention reasoning \cite{discern-span-generation} and dialogue graph reasoning \cite{graph-span-generation}, named \oursLongDiscern{} and \oursLong{} respectively. The parameters of entailment reasoning decoder are randomly initialized, the remain parameters are initialized with official T5 \cite{t5}. \oursLong{} and \oursLongDiscern{} are fine-tuned with AdamW \cite{loshchilov2018fixing-adamw} in 16 epochs, and the batch sizes are 32 and 16 respectively. We use hierarchical learning rates, the learning rate of T5 is 2e-4, the learning rate of other parameters are 2e-5. We've tried 1.0, 1.5, 2.0, 3.0 for $\lambda$, and find 1.0 is the best base on the results in the dev set. During inference decoding, the beam search number is set to 5. All results are conducted in two 3090 GPU (24GB memory)

\subsection{Results}

\begin{table*}[t]
\centering
\scalebox{0.99}{
\begin{tabular}{lcccc}
\toprule
\textbf{Models}& \textbf{Micro} & \textbf{Macro} & \textbf{BLEU-1} & \textbf{BLEU-4} \\ \midrule
Seq2Seq \cite{sharc} & 44.8 & 42.8 & 34.0 & 7.8 \\
Pipeline \cite{sharc} & 61.9 & 68.9 & 54.4 & 34.4 \\

BERTQA \cite{zhong-zettlemoyer-2019-e3-span-generation} & 63.6 & 70.8 & 46.2 & 36.3 \\

UrcaNet \cite{UrcaNet} & 65.1 & 71.2 & 60.5 & 46.1 \\

BiSon \cite{lawrence-etal-2019-bison-stage1} & 66.9 & 71.6 & 58.8 & 44.3 \\
$\mathrm{E^3}$ \cite{zhong-zettlemoyer-2019-e3-span-generation} & 67.6 & 73.3 & 54.1 & 38.7 \\
EMT \cite{emt-span-generation} & 69.1 & 74.6 & 63.9 & 49.5 \\

DISCERN \cite{discern-span-generation} & 73.2 & 78.3 & 64.0 & 49.1 \\
DGM \cite{graph-span-generation} & \textbf{77.4} & \textbf{81.2} & 63.3 & 48.4 \\
\midrule

\oursLongDiscern{} (ours) & \textbf{74.4} & \textbf{78.7} & \textbf{66.4} & \textbf{51.6} \\

\oursLong{} (ours)& 76.3 & 80.5 & \textbf{69.6} & \textbf{55.2} \\

\bottomrule
\end{tabular}
}
\caption{Performance on the blind held-out test set of ShARC benchmark.}
\label{table:testResult}
\end{table*}

\begin{table}[t]
\centering
\scalebox{0.73}{
\begin{tabular}{lccccc}
\toprule
\textbf{Models}& \textbf{Micro} & \textbf{Macro} &  \textbf{BLEU-1} & \textbf{BLEU-4} & \textbf{Params} \\

\midrule

Discern & 74.9 & \textbf{79.8} & 65.7 & 52.4 & 330M \\

\oursLongDiscern{} & \textbf{75.4} & 79.7 & 65.2 & 51.1 & 220M\\

\midrule

DGM & \textbf{78.6} & 82.2 & \textbf{71.8} & \textbf{60.2} & 1020M \\
\oursLong{} & \textbf{78.6} & \textbf{82.5}  & 65.3 & 53.3 & 770M\\

\bottomrule
\end{tabular}
}
\caption{Performance on the dev set of the ShARC benchmark. Params are the parameter numbers of PrLMs used in the framework.}
\label{table:devResult-para}
\end{table}


\begin{table}[t]
\centering
\scalebox{0.75}{
\begin{tabular}{lcccc}
\toprule
\multirow{2}{*}{\textbf{Models}}  & \multicolumn{2}{c}{\textbf{Dev Set}} & \multicolumn{2}{c}{\textbf{Test Set}}\\
 & \textbf{BLEU-1} & \textbf{BLEU-4}  & \textbf{BLEU-1}  & \textbf{BLEU-4}  \\ 
\midrule
$\mathrm{E^3}$ &  67.1 & 53.7 & 54.1(-13.0) & 38.7(-15.0) \\
EMT &  67.5 & 53.2 & 63.9(-3.6) & 49.5(-3.7) \\
DISCERN & 65.7 & 52.4 & 64.0(-1.7) & 49.1(-3.3) \\
DGM & 71.8 & 60.2 & 63.3(-8.5) & 48.4(-11.8) \\
\midrule
\oursLongDiscern{} & 65.2 & 51.1 & \textbf{66.4(+1.2)} & \textbf{51.6(+0.5)} \\
\oursLong{} & 65.3 & 53.3 & \textbf{69.6(+4.3)} & \textbf{55.2(+1.9)} \\

\bottomrule
\end{tabular}
}
\caption{Performance of BLEU scores on the dev set and test set of the ShARC benchmark.}
\label{table:consistentcy}
\end{table}

All results in the blind held-out test set of the ShARC benchmark are illustrated in Table \ref{table:testResult}. There are two different implementations here. \oursLongDiscern{} is configured with a DISCERN-formed entailment reasoning decoder by using the base-size model as the backbone. \oursLong{} is configured with a DGM-formed entailment reasoning decoder by using the large-size model as the backbone.

Experimental results demonstrate that the proposed framework achieves new SOTA with considerable improvement in terms of BLEU scores.  \oursLongDiscern{} outperforms DISCERN by 2.4 in BLEU-1, 2.5 in BLEU-4, 1.2 in micro-averaged accuracy, and 0.4 in macro-averaged accuracy. \oursLong{} outperforms DGM by 6.3 in BLEU-1, 6.8 in BLEU-4. We further analyze the results in the dev set shown in Table \ref{table:devResult-para}. Compared to the existing pipeline framework, our framework reduces the number of parameters by 32.5\% and 24.5\% for the base-size model and large-size model, respectively. 

Particularly, the BLEU scores of our \oursLong{} framework outperform DISCERN and DGM with a considerable improvement in the test set. Compared with the previous SOTA, the results have increased by 5.6 and 5.7 respectively in BLEU-1 and BLEU-4. Moreover, as shown in Table \ref{table:consistentcy}, the existing pipeline frameworks have a certain degree of decline on the test set with BLEU scores, which indicates the drawback of the existing pipeline architectures. In the contract, the BLEU scores of \oursLong{} and \oursLongDiscern{} continue to improve on the test set, which demonstrates the better generalization of our framework \oursLong{} in question generation. The above results prove that our proposed framework takes better advantage of the fine-grained entailment reasoning information and eliminate the information gap between decision making and question generation.


Additionally, in the decision making evaluation, we achieve the best performance in the dev set, but there is a slight drop in the test set. However, a good classification result must be an inference based on an existing fact. Intuitively, the correctness of reasoning can be analyzed by the performance of the question generation. Correct reasoning will make the model ask the right questions. Correct classification, but asking the wrong question, does not mean that the model has learned the reasoning ability correctly, and the phenomena such as statistical bias may also cause this problem.




 







\subsection{Ablation Studies}

The existing generation question evaluation metrics suffer from randomness\footnote{ The generated questions are evaluated with BLEU scores only if the prediction and ground truth in classification are both 'inquire'.} on the small dev set (2,270). To better evaluate the question generation abilities of models on the dev set, we utilized ALL-BLEU (ABLEU) to evaluate all the examples that ground truth is inquired to generate a question in ablation studies. All the other settings remain the same with official evaluation.

The ablation studies of \oursLong{} on the dev set on ShARC benchmark are shown in Table \ref{table:devAblation-base}. We use the base-size model to investigate the impacts of different components, there are three ablations of our \oursLong{}-Base is considered:

\begin{itemize}
\item \textbf{\oursLong{}-Base-wo/g} trains the model without graph reasoning block, the setting is the same as \oursLongDiscern{}.

\item \textbf{\oursLong{}-Base-wo/g+f} trains the model without graph reasoning block and fine-grained prefix.

\item \textbf{\oursLong{}-Base-wo/e+f} trains the model without entailment reasoning decoder and fine-grained prefix, which can be considered as the original T5 model.
\end{itemize}


\subsubsection{Analysis of Graph Reasoning}
Graph Reasoning consists of explicit discourse graph reasoning and implicit discourse graph reasoning, each of them introducing discourse relations among EDUs and decoupling-fusion mechanism into \oursLong{}, respectively. This setting is the same as \oursLongDiscern{}. Both accuracy scores and ABLEU scores are improved by introducing graph reasoning. In addition, we observe a significant reduction in the ABLEU scores if removing graph reasoning. ABLEU is used to measure whether the model answers due to the correct reasoning of the missing knowledge, the results show \oursLong{}-Base correctly reasoned out the missing knowledge, which suggests the necessity of graph reasoning block.

\begin{table}[t]
\centering
\scalebox{0.75}{
\begin{tabular}{lcccc}
\toprule
\textbf{Models}& \textbf{Micro} & \textbf{Macro} & \textbf{ABLEU-1} & \textbf{ABLEU-4} \\ \midrule
\oursLong{}-Base & \textbf{75.9} & \textbf{80.4} & \textbf{54.7} & \textbf{43.6}    \\

\oursLong{}-Base-wo/g & 75.4 & 79.7 & 45.0& 36.4\\

\oursLong{}-Base-wo/g+f  & 73.4 & 78.0 & 49.4 & 40.3 \\

\oursLong{}-Base-wo/e+f & 72.9 & 77.3 & 42.1 & 35.0  \\

\bottomrule
\end{tabular}
}
\caption{Ablation study of our base-size model on the dev set of ShARC.}
\label{table:devAblation-base}
\end{table}

\subsubsection{Analysis of Fine-grained Text Prefix}
 We investigate the necessity of the fine-grained text prefix by additional removing the fine-grained text prefix in \oursLongDiscern{}, while it's hard to reason for the entailment of EDUs without the fine-grained special prefix. We feed fine-grained special tokens prefixed text into \oursLong{} directly. As shown in the results, compared with \oursLong{}-Base-wo/g and \oursLong{}-Base-wo/g+f, the accuracy will be significantly improved by introducing the fine-grained text prefix, which indicates that directly using special token prefixes will cause noise disturbance for semantic learning. As illustrated in \ref{table:devAblation-base}, the ABLEU-1 is decreased by 4.4, and the ABLEU-4 is decreased by 3.9. The above results show the importance of the fine-grained text prefix.

\subsubsection{Analysis of Entailment Reasoning}
\oursLong{}-Base-wo/e+f can be considered as the official T5 model. As shown in Table \ref{table:devAblation-base}, the lack of fine-grained entailment reasoning information will seriously affect the performance of decision making and question generation. Compared with the performance of \oursLong{}-Base, the ABLEU-1 and ABLEU-4 of \oursLong{}-Base-wo/e+f decreased by 7.3 and 5.3 after removing entailment reasoning decoder, which indicates the importance of entailment reasoning, especially for reasoning of question generation. 

\section{Conclusion}

In this paper, we propose a novel end-to-end framework, called \oursLong{}, to better capture the entailment information for question generation in CMRC, and thus eliminate the information gap between decision making and question generation. By conducting a parameter shared encoder between answer generation decoder and entailment reasoning decoder, the answer generation decoder can utilize the fine-grained entailment reasoning information to enhance the performance of question generation. Experimental results suggest that the proposed framework \oursLong{} achieves the new state-of-the-art results on the ShARC benchmark. 

\section*{Acknowledgements}
 The work is supported by National Key R\&D Plan (No.2020AAA0106600), National Natural Science Foundation of China (No.U21B2009, 62172039 and L1924068).
\balance
\bibliographystyle{acl_natbib}
\bibliography{cc}

\newpage
\appendix


\end{document}